\documentclass[lettersize,journal]{IEEEtran}
\usepackage{amsmath,amsfonts}
\usepackage{algorithmic}
\usepackage[linesnumbered,ruled]{algorithm2e}
\SetKwInOut{Input}{Input}
\SetKwInOut{Output}{Output}
\usepackage{array}
\usepackage[caption=false,font=normalsize,labelfont=sf,textfont=sf]{subfig}
\usepackage{textcomp}
\usepackage{stfloats}
\usepackage{url}
\usepackage{verbatim}
\usepackage{graphicx}
\usepackage{cite}
\usepackage{xcolor}
\usepackage{booktabs}
\begin{document}

\title{Spatially Adaptive Frequency Conditioning for Fourier-Encoded Implicit Neural Representations}

\author{Ligen Shi, Jun Qiu, Zengyu Pang, Yuhang Zheng, Lina Wu, Di He, and Chang Liu%
\thanks{Ligen Shi is with the College of Computer Science (College of Software), Inner Mongolia University, Hohhot 010021, China
(e-mail: ligenshi0826@gmail.com).}%
\thanks{Jun Qiu, Yuhang Zheng, Lina Wu, Di He, and Chang Liu are with the Institute of Computational Imaging, Beijing Information Science and Technology University, Beijing 102206, China(e-mail: qiujun@bistu.edu.cn; zhengyuhang206@163.com;
wulina@bistu.edu.cn; dihe@bistu.edu.cn; changliuct@gmail.com).}%
\thanks{Zengyu Pang is with SKLMS, Academy of Mathematics and Systems Science, Chinese Academy of Sciences, Beijing 100190, China (e-mail: pbfpzy@amss.ac.cn).}}

\markboth{Journal of \LaTeX\ Class Files,~Vol.~, No.~, ~2026}%
{Ligen Shi \MakeLowercase{\textit{et al.}}: Spatially Adaptive Frequency Conditioning for Fourier-Encoded INR}

\IEEEpubid{0000--0000/00\$00.00~\copyright~2024 IEEE}

\maketitle

\begin{abstract}
Implicit neural representations (INRs) commonly employ Fourier features to alleviate spectral bias and represent high-frequency signals. Fourier encodings, however, use the same frequency representation across spatial locations despite the spatially varying frequency content of natural signals. We propose Adaptive Frequency Implicit Neural Representation (AFINR), a spatially adaptive frequency-conditioning framework for Fourier-encoded INRs. A sampling-compatible Fourier dictionary first defines the explicit frequency support, and a learnable spatial control field then modulates the admitted Fourier components before decoding. The control field is parameterized in the discrete cosine transform (DCT) domain to provide a compact and continuously queryable spatial representation. AFINR can be combined with different coordinate decoders, including ReLU, SIREN, CAFE, and CAFE+. Experiments on Kodak and DIV2K show consistent average improvements across these decoder families. In particular, AFINR-CAFE+ improves CAFE+ by 3.95 dB on Kodak and 3.50 dB on DIV2K. Ablation studies characterize the effects of adaptive-field parameterization, DCT coefficient retention, and frequency-response parameters, while learned spatial responses show position-dependent use of the admitted Fourier components.
\end{abstract}

\begin{IEEEkeywords}
Implicit neural representation, Fourier features, positional encoding, adaptive frequency filtering, continuous representation, spectral bias.
\end{IEEEkeywords}

\section{Introduction}
\label{sec:introduction}
\IEEEPARstart{I}{mplicit} neural representations (INRs), also known as neural fields, represent signals as continuous functions of coordinates.
They have been widely used for geometry representation ~\cite{park2019deepsdf,mescheder2019occupancy,chen2019imnet},
neural scene modeling and novel-view synthesis ~\cite{sitzmann2019srn,mildenhall2020nerf},
continuous image representation~\cite{chen2021liif,lee2022lte},
neural compression~\cite{dupont2021coin,dupont2022coinpp}, and
computational imaging~\cite{shi2025ndf,liu2026raylf,shi2025ringinr}.
Recent surveys summarize advances in activation functions, positional
encoding, and network design~\cite{xie2022neuralfields,essakine2025survey}. Despite these developments, the expressive capacity of coordinate MLPs
for high-frequency and locally complex signals remains limited, motivating richer frequency representations ~\cite{rahaman2019spectral,xu2019frequency,basri2020frequency}.

Existing INR methods improve the representation of high-frequency and
locally complex signals from several directions. One line modifies the
nonlinear activation of the coordinate network. SIREN~\cite{sitzmann2020siren}
uses periodic sinusoidal activations, while subsequent studies analyze
their spectral properties~\cite{ramasinghe2022beyond}. WIRE
~\cite{saragadam2023wire}, FINER~\cite{liu2024finer}, SL$^2$A-INR
~\cite{rezaeian2025sl2a}, MIRE~\cite{jayasundara2025mire}, and SAFIR
~\cite{zeraatkar2026safir} further improve frequency representation
through localized wavelets, adaptive activations, or signal-dependent
frequency modulation. These methods primarily modify the spectral
characteristics of the nonlinear network rather than the spatial use
of an explicit Fourier encoding.

\IEEEpubidadjcol

A second line focuses on frequency representation and input encoding.
Random Fourier features~\cite{rahimi2007random} and Fourier feature
mappings~\cite{tancik2020fourier} enrich the input frequency space,
while positional encoding has become standard in neural fields
~\cite{mildenhall2020nerf}. Sampling-aware encodings such as mip-NeRF
~\cite{barron2021mipnerf,barron2022mipnerf360}, multiresolution hash
encoding~\cite{muller2022instantngp}, and progressive frequency
activation~\cite{hertz2021sape,lin2021barf,yang2023freenerf} further
improve frequency control. CAFE~\cite{ke2026cafe} instead uses
multiplicative Fourier interactions, together with Chebyshev features
in CAFE+, to synthesize richer frequencies before decoding.

A third family modifies the architecture or parameterization of INRs.
MFN~\cite{fathony2021mfn} and BACON~\cite{lindell2022bacon} introduce
filter-based or band-limited structures; ACORN~\cite{martel2021acorn}
and MINER~\cite{saragadam2022miner} allocate capacity across spatial
regions or scales; and DINER~\cite{xie2023diner} and
Fourier-reparameterized training~\cite{shi2024frinr} modify coordinate
or network parameterization to improve signal fitting.

\begin{figure}[t]
\centering
\includegraphics[width=0.90\columnwidth]{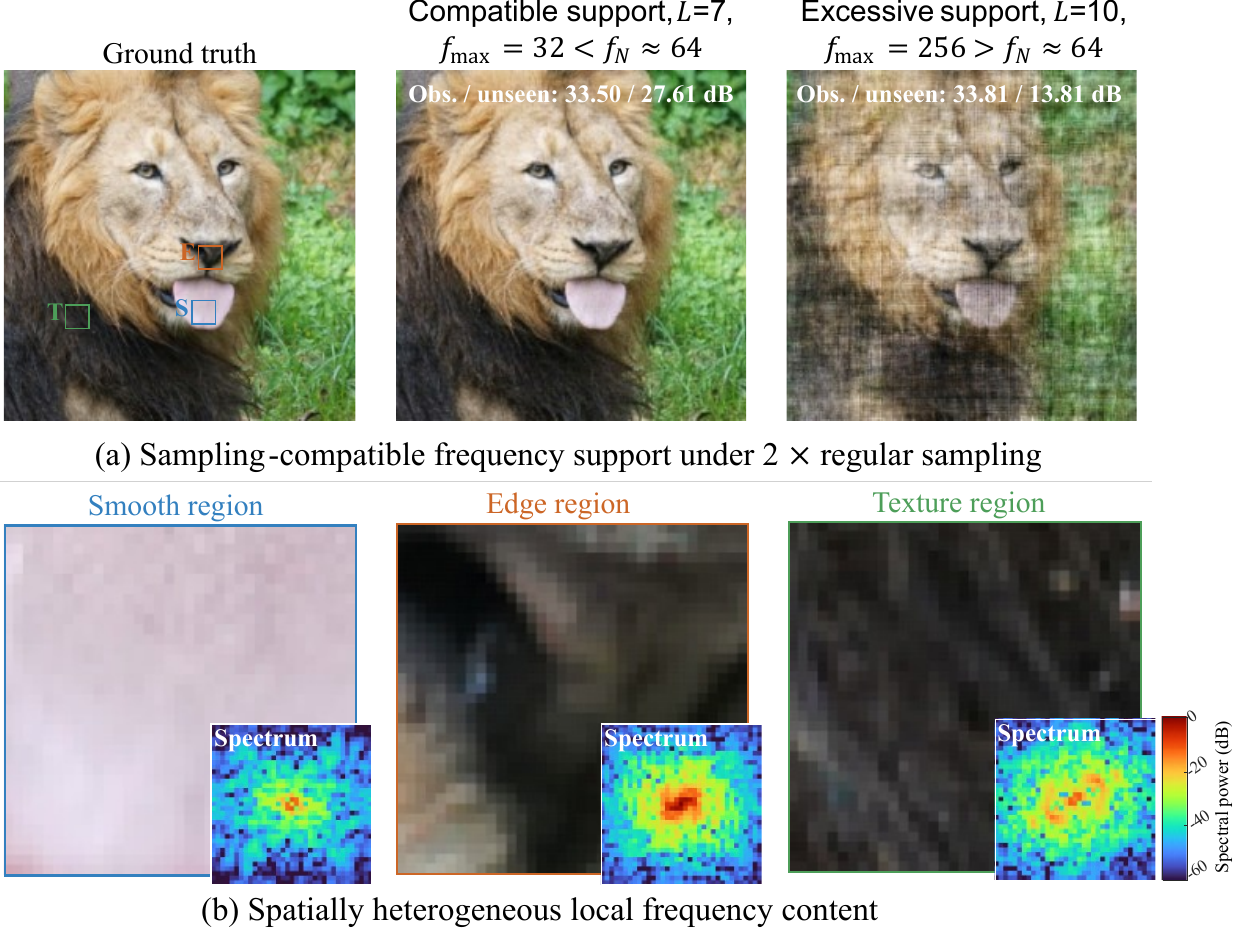}
\caption{Motivation for frequency control in Fourier-encoded INRs. (a) Under $2\times$ regular sampling, $L=7$ remains within the sampling-compatible range ($f_{\max}=32<f_{\mathrm{N}}\approx64$), whereas $L=10$ exceeds it ($f_{\max}=256>f_{\mathrm{N}}\approx64$), improving fitting at observed coordinates but degrading reconstruction at unseen coordinates. (b) Equal-sized smooth, edge, and texture regions exhibit different local spectra, computed using the same Hann-windowed Fourier analysis and common spectral-power reference.}
\label{fig:motivation}
\end{figure}

As illustrated in Fig.~\ref{fig:motivation}, we identify two issues in frequency control for Fourier-encoded INRs: the explicit frequency support should be compatible with the available sampling density, while the effective use of the admitted frequencies should adapt to spatially varying signal content.

The first issue concerns frequency support under finite sampling. An INR is optimized from discrete observations but represents a continuous signal. Classical sampling theory relates recoverable frequency content to the sampling density ~\cite{shannon1949}. Although this does not impose a strict band limit on the output of a nonlinear INR, the explicit Fourier features introduce controllable oscillatory components. When their frequency support substantially exceeds what can be constrained by the observations, fitting at sampled coordinates may improve while reconstruction between them deteriorates, as illustrated in Fig.~\ref{fig:motivation}(a). Related neural-field studies have also highlighted the importance of frequency restriction, integrated sampling, and antialiasing ~\cite{barron2021mipnerf,lindell2022bacon,barron2022mipnerf360,
barron2023zipnerf,yang2023freenerf}. The explicit Fourier support should therefore be selected according to the sampling resolution.

The second issue arises after an appropriate global frequency range has been selected. Conventional Fourier encoding supplies the same frequency dictionary to every spatial location, although local frequency content varies across smooth regions, edges, fine structures, and textures, as shown in Fig.~\ref{fig:motivation}(b).

Related methods address frequency representation from different perspectives. SAPE~\cite{hertz2021sape} adapts the progression of positional encoding, SAFIR~\cite{zeraatkar2026safir} predicts coordinate-dependent sinusoidal frequencies, SASNet~\cite{feng2026sasnet} uses learned spatial masks, and SCONE~\cite{li2024scone} dispatches different Fourier bases to different regions. CAFE~\cite{ke2026cafe} enhances frequency composition through multiplicative interactions. FA-INR~\cite{zhou2025fainr} and CF-INR~\cite{yu2026cfinr} further explore explicit frequency modeling from different perspectives. FA-INR introduces an additional frequency dimension and represents multidimensional signals in a learnable latent frequency space, where different frequency components are decoupled and subsequently integrated into the original signal domain. CF-INR performs frequency-wise representation in the Haar wavelet domain and models relationships among wavelet components through cross-frequency tensor decomposition with self-evolving parameters. These methods focus on the explicit representation and interaction of different frequency components. In contrast, AFINR is motivated by spatial variation in signal frequency content. It keeps the sampling-compatible Fourier dictionary fixed and uses a scalar spatial control field to generate a structured response over the ordered frequency axis, thereby modulating the contribution of the admitted Fourier components across spatial locations.

Motivated by these observations, we propose Adaptive Frequency Implicit Neural Representation (AFINR), which separates global frequency support from spatial frequency adaptation. A sampling-compatible Fourier dictionary first determines the explicit frequencies available under the given sampling condition, while a learnable spatial field controls how the admitted Fourier atoms contribute at different locations before decoding. Unlike methods that modify sinusoidal activation frequencies~\cite{sitzmann2020siren,liu2024finer,zeraatkar2026safir}, allocate model capacity across spatial regions
~\cite{martel2021acorn,saragadam2022miner}, synthesize additional frequencies through feature interactions~\cite{ke2026cafe}, or represent
different frequency components separately~\cite{zhou2025fainr,yu2026cfinr}, AFINR keeps the admitted Fourier dictionary fixed and adapts its effective use across spatial locations.

The two levels of frequency control have distinct roles. Global frequency support determines which explicit frequencies are available
under a given sampling condition, whereas spatial frequency conditioning determines how the admitted frequencies participate at each location. The sampling-compatible support therefore provides the global Fourier basis on which the proposed spatial conditioning operates. Our experimental evaluation focuses on the spatial adaptation mechanism, its compatibility with different decoder architectures, and the representation of the adaptive field.

The main contributions of this work are summarized as follows.
\begin{itemize}
    \item We formulate frequency control in Fourier-encoded INRs at
    two complementary levels: a sampling-compatible global Fourier
    support determines the explicit frequencies available to the
    representation, while spatial frequency conditioning controls
    how the admitted frequencies are used at each location.

    \item We develop a spatially adaptive Fourier representation in
    which a learnable position-dependent response modulates the
    admitted Fourier atoms. A DCT-parameterized control field provides
    a compact and continuously queryable representation of the spatial
    adaptation.
\end{itemize}
Experiments with ReLU, SIREN, CAFE, and CAFE+ demonstrate consistent
improvements and analyze the proposed frequency-conditioning mechanism.

\section{Method}
\label{sec:method}
\subsection{Problem Formulation}
\label{sec:problem_formulation}

\begin{figure*}[t]
    \centering
    \includegraphics[width=0.92\textwidth]{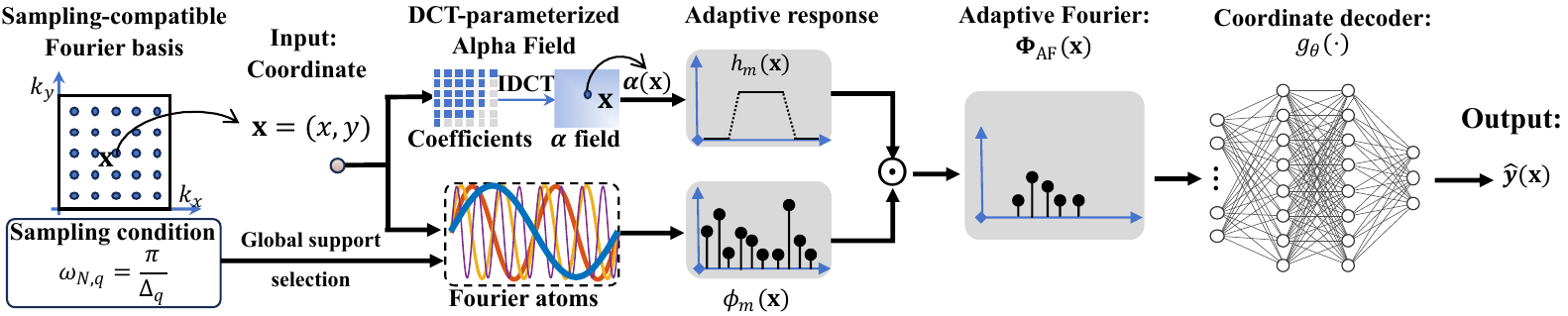}
    \caption{Overview of AFINR. The sampling condition defines the Fourier basis $\mathcal{W}_{\mathrm{s}}$, while the spatial control field $\alpha(\mathbf{x})$ generates responses $h_m(\mathbf{x})$ that modulate the admitted Fourier atoms. The resulting adaptive representation $\boldsymbol{\Phi}_{\mathrm{AF}}(\mathbf{x})$ is mapped to $\hat{\mathbf{y}}(\mathbf{x})$ by the coordinate decoder $g_{\theta}$.}
    \label{fig:framework}
\end{figure*}

Given $N$ observations $\mathcal{D}=\{(\mathbf{x}_n,\mathbf{y}_n)\}_{n=1}^{N}$, where $\mathbf{x}_n\in\Omega\subset\mathbb{R}^{d_{\mathrm{in}}}$ denotes a coordinate in the signal domain $\Omega$ and $\mathbf{y}_n\in\mathbb{R}^{d_{\mathrm{out}}}$ is the corresponding
signal value, we seek a continuous coordinate-based representation
of the underlying signal.

For the $m$-th Fourier frequency vector $\boldsymbol{\omega}_m\in\mathbb{R}^{d_{\mathrm{in}}}$, the corresponding sine--cosine atom is defined as
\begin{equation}
    \boldsymbol{\phi}_m(\mathbf{x})
    =
    \begin{bmatrix}
        \sin(\boldsymbol{\omega}_m^{\top}\mathbf{x})\\
        \cos(\boldsymbol{\omega}_m^{\top}\mathbf{x})
    \end{bmatrix}.
    \label{eq:fourier_atom}
\end{equation}

Let $\mathcal{W}_{\mathrm{s}} =\{\boldsymbol{\omega}_m\}_{m=1}^{M_{\mathrm{s}}}$ denote the sampling-compatible Fourier dictionary, where $M_{\mathrm{s}}=|\mathcal{W}_{\mathrm{s}}|$ is the number of Fourier atoms admitted by the sampling condition, as detailed in Sec.~\ref{sec:sampling_support}. For each admitted atom, $h_m(\mathbf{x})\in\mathbb{R}$ denotes its position-dependent response at coordinate $\mathbf{x}$.

The adaptive Fourier representation is then defined by
\begin{equation}
    \boldsymbol{\Phi}_{\mathrm{AF}}(\mathbf{x})
    =
    \left[
        h_m(\mathbf{x})\boldsymbol{\phi}_m(\mathbf{x})
    \right]_{m=1}^{M_{\mathrm{s}}},
    \label{eq:adaptive_fourier_representation}
\end{equation}
where $[\cdot]_{m=1}^{M_{\mathrm{s}}}$ denotes concatenation over the admitted Fourier atoms. The signal is predicted as
\begin{equation}
    \hat{\mathbf{y}}(\mathbf{x})
    =
    g_{\theta}\!\left(
        \boldsymbol{\Phi}_{\mathrm{AF}}(\mathbf{x})
    \right),
    \label{eq:afinr_model}
\end{equation}
where $g_{\theta}:\mathbb{R}^{2M_{\mathrm{s}}} \rightarrow\mathbb{R}^{d_{\mathrm{out}}}$ is a coordinate decoder with learnable parameters $\theta$.

The overall AFINR framework is illustrated in Fig.~\ref{fig:framework}, where the sampling-compatible Fourier dictionary provides the global frequency support and the spatial control field generates position-dependent responses for the admitted Fourier atoms.

\subsection{Sampling-Compatible Fourier Basis}
\label{sec:sampling_support}

We construct a sampling-compatible Fourier dictionary by retaining only the frequencies supported by the sampling condition. Let $\mathcal{W}=\{\boldsymbol{\omega}_m\}_{m=1}^{M}$ denote a candidate Fourier dictionary.

For the commonly used dyadic positional encoding, the frequency vector along the $q$th coordinate dimension is
\begin{equation}
    \boldsymbol{\omega}_{j,q}
    =
    2^{j}\pi\mathbf{e}_{q},
    \qquad
    j=0,\ldots,L-1,
    \label{eq:dyadic_frequency}
\end{equation}
where $\mathbf{e}_{q}$ is the $q$th standard basis vector. The formulation is not restricted to this dyadic construction and can be applied to general multidimensional Fourier dictionaries.

For regularly sampled coordinates, let $\Delta_q$ denote the sampling interval along dimension $q$. The corresponding Nyquist angular frequency is
\begin{equation}
    \omega_{\mathrm{N},q}
    =
    \frac{\pi}{\Delta_q}.
    \label{eq:nyquist_frequency}
\end{equation}

We retain only candidate frequency vectors whose components lie within the sampling-supported range,
\begin{equation}
    \mathcal{W}_{\mathrm{s}}
    =
    \left\{
        \boldsymbol{\omega}_{m}\in\mathcal{W}
        \;\middle|\;
        |\omega_{m,q}|
        \leq
        \omega_{\mathrm{N},q},
        \quad
        q=1,\ldots,d_{\mathrm{in}}
    \right\}.
    \label{eq:sampling_support}
\end{equation}
The resulting dictionary contains
$M_{\mathrm{s}}=|\mathcal{W}_{\mathrm{s}}|$ admitted Fourier atoms.

Eq.~\eqref{eq:sampling_support} constrains only the frequencies explicitly supplied by the Fourier dictionary; it does not impose a strict band limit on the network output, since spatial modulation and nonlinear decoding may generate additional spectral components. Under finite optimization, however, the spectral bias of neural networks makes such higher-frequency components generally more difficult to learn. If all candidate frequencies satisfy the
sampling condition, $\mathcal{W}_{\mathrm{s}}=\mathcal{W}$.

\subsection{Spatially Adaptive Fourier Representation}
\label{sec:adaptive_fourier}

Given the fixed sampling-compatible dictionary $\mathcal{W}_{\mathrm{s}}$, AFINR assigns a position-dependent response to each admitted Fourier atom.

\subsubsection{Spatial Frequency Response}
\label{sec:spatial_frequency_response}

To define the response over the ordered frequency axis, let $\rho_m=\|\boldsymbol{\omega}_m\|_2$ denote the radial magnitude of the $m$th frequency vector. Let $\rho^{(0)}<\rho^{(1)}<\cdots<\rho^{(K-1)}$
denote the $K$ distinct magnitudes in $\mathcal{W}_{\mathrm{s}}$. If $\rho_m=\rho^{(k_m)}$, we define its normalized frequency coordinate as
\begin{equation}
    r_m =
    \begin{cases}
        \displaystyle\frac{k_m}{K-1}, & K>1,\\[4pt]
        0, & K=1,
    \end{cases}
    \qquad r_m\in[0,1].
    \label{eq:normalized_frequency_rank}
\end{equation}
Here, $r_m$ denotes the normalized rank of the frequency magnitude. Sine and cosine components associated with the same $\boldsymbol{\omega}_m$ share the same $r_m$ and response.

We introduce a scalar spatial control field $\alpha:\Omega\rightarrow\mathbb{R}$. For the $m$th admitted Fourier atom, its response at $\mathbf{x}$ is
\begin{equation}
\begin{split}
    h_m(\mathbf{x})
    =&\;
    \sigma\!\left[
        \kappa
        \left(
            r_m-\alpha(\mathbf{x})+\frac{B}{2}
        \right)
    \right] \\
    &-
    \sigma\!\left[
        \kappa
        \left(
            r_m-\alpha(\mathbf{x})-\frac{B}{2}
        \right)
    \right],
    \label{eq:adaptive_response}
\end{split}
\end{equation}
where $\sigma(z)=1/(1+\exp(-z))$, $B$ controls the response width, and $\kappa$ controls the transition steepness. Changing $\alpha(\mathbf{x})$ shifts this smooth band-shaped response along the normalized frequency axis, allowing the contribution of the fixed Fourier atoms to vary with spatial location.

Substituting Eq.~\eqref{eq:adaptive_response} into Eq.~\eqref{eq:adaptive_fourier_representation} yields
\begin{equation}
    \boldsymbol{\Phi}_{\mathrm{AF}}(\mathbf{x})
    =
    \begin{bmatrix}
        h_1(\mathbf{x})
        \sin(\boldsymbol{\omega}_1^{\top}\mathbf{x})
        \\
        h_1(\mathbf{x})
        \cos(\boldsymbol{\omega}_1^{\top}\mathbf{x})
        \\
        \vdots
        \\
        h_{M_{\mathrm{s}}}(\mathbf{x})
        \sin(\boldsymbol{\omega}_{M_{\mathrm{s}}}^{\top}\mathbf{x})
        \\
        h_{M_{\mathrm{s}}}(\mathbf{x})
        \cos(\boldsymbol{\omega}_{M_{\mathrm{s}}}^{\top}\mathbf{x})
    \end{bmatrix}.
    \label{eq:adaptive_fourier_expanded}
\end{equation}

AFINR therefore retains a fixed sampling-compatible Fourier dictionary while making the contribution of each Fourier atom spatially dependent through $\alpha(\mathbf{x})$.

\subsubsection{DCT-Parameterized Spatial Frequency Field}
\label{sec:dct_alpha_field}

To compactly represent the spatial control field, we parameterize $\alpha(\mathbf{x})$ in the discrete cosine transform (DCT) domain.

For a two-dimensional signal of size $H\times W$, we retain $K_{\alpha}=\operatorname{round}(r_{\mathrm{DCT}}HW)$ DCT coefficients, where $r_{\mathrm{DCT}}\in(0,1]$ is the retention ratio. The DCT indices $(u,v)$ are ordered according to their radial index $d_{u,v}=\sqrt{u^2+v^2}$, and the first $K_{\alpha}$ indices form
the retained set $\mathcal{S}_{K_{\alpha}}$. Let $\mathbf{c}_{\alpha}\in\mathbb{R}^{K_{\alpha}}$ denote the learnable coefficients, and let $\mathbf{C}_{\alpha}\in\mathbb{R}^{H\times W}$ denote the corresponding DCT coefficient matrix, with all unretained entries
set to zero.

The spatial control field and its continuous query are obtained as
\begin{equation}
    \mathbf{A}_{\alpha}
    =
    \operatorname{IDCT}_{2}(\mathbf{C}_{\alpha}),
    \qquad
    \alpha(\mathbf{x})
    =
    \mathcal{I}(\mathbf{x};\mathbf{A}_{\alpha}),
    \label{eq:dct_alpha}
\end{equation}
where $\mathcal{I}$ denotes bilinear interpolation. The retention ratio $r_{\mathrm{DCT}}$ controls the accuracy--compactness tradeoff of the adaptive field. We use $r_{\mathrm{DCT}}=0.60$ by default and examine its effect in Sec.~\ref{sec:ablation_dct_ratio}.

AFINR also includes two limiting cases: setting $\alpha(\mathbf{x})=\alpha_{\mathrm{g}}$ yields a globally shared frequency response, while setting $h_m(\mathbf{x})\equiv1$ recovers the conventional unmodulated Fourier representation.

\subsection{Learning the Continuous Representation}
\label{sec:learning}

The decoder parameters $\theta$ and the DCT coefficients $\mathbf{c}_{\alpha}$ are jointly optimized by minimizing the reconstruction error over the observed coordinates,
\begin{equation}
    \mathcal{L}
    =
    \frac{1}{N}
    \sum_{n=1}^{N}
    \left\|
        g_{\theta}
        \left(
            \boldsymbol{\Phi}_{\mathrm{AF}}(\mathbf{x}_n)
        \right)
        -
        \mathbf{y}_n
    \right\|_2^2.
    \label{eq:fitting_loss}
\end{equation}
During optimization, the sampling-compatible dictionary $\mathcal{W}_{\mathrm{s}}$ remains fixed, while $\theta$ and $\mathbf{c}_{\alpha}$ are learned jointly. No direct supervision is imposed on $\alpha(\mathbf{x})$ or $h_m(\mathbf{x})$.

For $N$ queried coordinates and $M_{\mathrm{s}}$ admitted Fourier atoms, evaluating the adaptive Fourier representation requires $\mathcal{O}(NM_{\mathrm{s}})$ coordinate-dependent operations. The DCT parameterization introduces $K_{\alpha}$ additional learnable coefficients without modifying the coordinate decoder.

\section{Experiments}\label{sec:experiments}
We evaluate AFINR on dense natural-image fitting through comparisons with representative INR methods, controlled decoder-wise studies, ablations, and analyses of parameter efficiency and learned spatial frequency adaptation.

\subsection{Experimental Setup}
\label{sec:experimental_setup}

\textbf{Datasets and tasks.}
We evaluate AFINR on two natural-image benchmarks, Kodak~\cite{kodak1993} and DIV2K~\cite{agustsson2017div2k}. All 24 Kodak images are used at their native resolution, i.e., $768\times512$ or $512\times768$ depending on image orientation. For DIV2K, we use validation images 0801--0824 and extract a deterministic $512\times512$ center crop from each image. RGB intensities are normalized to $[0,1]$, and the two-dimensional coordinates are normalized to $[-1,1]^2$. We consider dense image fitting, where all pixels are used as observations and evaluation is performed on the complete image grid.

\textit{Compared methods.}
We compare AFINR with representative INR methods covering different frequency-representation strategies, including PE-MLP~\cite{tancik2020fourier}, SIREN~\cite{sitzmann2020siren}, WIRE~\cite{saragadam2023wire}, FINER~\cite{liu2024finer}, SL$^2$A-INR~\cite{rezaeian2025sl2a}, CAFE and CAFE+~\cite{ke2026cafe}, SASNet~\cite{feng2026sasnet}, SAFIR~\cite{zeraatkar2026safir}, SCONE~\cite{li2024scone}, and CF-INR~\cite{yu2026cfinr}. We additionally construct PE-SIREN as a controlled baseline for AFINR-SIREN. PE-SIREN uses the same sampling-compatible Fourier dictionary and SIREN decoder as AFINR-SIREN, while disabling the spatially adaptive response ($h_m(\mathbf{x})\equiv1$).

Whenever an official implementation is available, we follow the recommended network configuration, initialization, and method-specific hyperparameters. For the baseline--AFINR comparisons, the decoder architecture and optimization budget are kept fixed. Parameter counts are reported separately to account for the additional adaptive field introduced by AFINR.

\textbf{Evaluation metrics.}
Reconstruction quality is evaluated using peak signal-to-noise ratio (PSNR), structural similarity index (SSIM), and learned perceptual image patch similarity (LPIPS). Higher PSNR and SSIM and lower LPIPS indicate better reconstruction quality. Unless otherwise specified, these metrics are computed on the complete reconstructed image. We additionally report the number of learnable parameters, training time, and peak GPU memory to characterize computational cost.

\textbf{Implementation details.}
All models are optimized using the mean squared error between the predicted and observed RGB values. Unless otherwise specified, each model is trained for 20,000 optimization steps, with checkpoints at 1,000, 5,000, 10,000,
and 20,000 steps retained for convergence analysis. We use Adam as the optimizer and sample 65,536 coordinates per optimization step when minibatch training is required; methods with a native full-grid formulation, such as CF-INR, retain their corresponding coordinate layout. Method-specific learning rates and initialization schemes follow their recommended settings, while the DCT-ratio and frequency-response ablations use the same decoder and optimization configuration. The AFINR-specific parameters, including the response width $B$, transition sharpness $\kappa$, and DCT coefficient retention ratio $r_{\mathrm{DCT}}$, are fixed across test images unless explicitly varied in the corresponding ablation study.

All experiments were implemented in Python 3.12 with PyTorch 2.8.0 and CUDA 12.8 on Ubuntu 22.04. The reported runtime and memory measurements were obtained from single-GPU runs using NVIDIA GeForce RTX 3080 Ti GPUs with 12 GB memory.

\subsubsection{Network Configurations}
\label{sec:network_configurations}

To evaluate decoder compatibility, we instantiate AFINR with multiple INR backbones. Unless otherwise specified, all variants use the DCT-parameterized adaptive field with $r_{\mathrm{DCT}}=0.60$, $B=0.50$, and $\kappa=12$. The dyadic frequency levels are determined by the sampling condition. For $N_q$ endpoint-inclusive
samples on $[-1,1]$, $\Delta_q=2/(N_q-1)$ and
\begin{equation}
    L_q =
    1+\left\lfloor
    \log_2\frac{N_q-1}{2}
    \right\rfloor .
\end{equation}
Thus, $512\times512$ images use $L_x=L_y=8$ ($M_{\mathrm{s}}=16$), while $768\times512$ images use $L_x=9$ and $L_y=8$, respectively.

AFINR is instantiated with ReLU, SIREN, CAFE, and CAFE+ decoders. For AFINR-ReLU, we use a four-layer MLP with a hidden width of 256, ReLU activations, and a sigmoid RGB output. AFINR-SIREN uses four hidden layers of width 256 with $\omega_0=30$. For AFINR-CAFE, the conditioned Fourier features are processed by the CAFE multiplicative decoder using 144 Fourier features with a scale factor of 30, a hidden width of 288, three multiplicative branches, and two backbone layers; no Chebyshev branch is used. AFINR-CAFE+ retains the original CAFE+ dual-branch structure, with adaptive conditioning applied only to the Fourier branch and a Chebyshev order of 32; the remaining decoder settings are identical to those of AFINR-CAFE.

PE-MLP and PE-SIREN provide fixed-Fourier references for the ReLU and SIREN variants, respectively, while CAFE and CAFE+ retain their original frequency-processing structures. AFINR-ReLU and AFINR-SIREN use a learning rate of $1\times10^{-4}$; the CAFE-based variants follow their corresponding CAFE optimization settings.

\subsection{Comparison with Representative INRs}

\begin{table*}[!h]
\centering
\caption{Quantitative comparison on Kodak and DIV2K at 20,000 optimizer updates. Results are mean $\pm$ standard deviation across test images. Params are reported at $512\times512$; Train is averaged over the 24 DIV2K images, and Mem. denotes peak GPU memory.}
\label{tab:main_comparison}
\scriptsize
\setlength{\tabcolsep}{2.2pt}
\begin{tabular}{lccccccccc}
\toprule
Method & \multicolumn{3}{c}{Kodak} & \multicolumn{3}{c}{DIV2K} & Params (M)$\downarrow$ & Train (min)$\downarrow$ & Mem. (GB)$\downarrow$ \\
\cmidrule(lr){2-4}\cmidrule(lr){5-7}
 & PSNR(dB)$\uparrow$ & SSIM$\uparrow$ & LPIPS$\downarrow$ & PSNR(dB)$\uparrow$ & SSIM$\uparrow$ & LPIPS$\downarrow$ & & & \\
\midrule
PE-MLP & $31.05 \pm 2.92$ & $0.8246 \pm 0.0509$ & $0.1921 \pm 0.0544$ & $32.62 \pm 3.45$ & $0.8834 \pm 0.0611$ & $0.1165 \pm 0.0439$ & 0.209 & 4.28 & 0.42 \\
SIREN & $32.78 \pm 3.48$ & $0.8558 \pm 0.0532$ & $0.2437 \pm 0.0763$ & $33.75 \pm 4.33$ & $0.9163 \pm 0.0609$ & $0.1191 \pm 0.0615$ & 0.199 & 4.13 & 0.53 \\
PE-SIREN & $34.76 \pm 2.18$ & $0.9605 \pm 0.0103$ & $0.0259 \pm 0.0111$ & $34.84 \pm 1.79$ & $0.9753 \pm 0.0139$ & $0.0151 \pm 0.0069$ & 0.207 & 3.56 & 0.67 \\
FINER & $38.54 \pm 3.06$ & $0.9608 \pm 0.0111$ & $0.0262 \pm 0.0124$ & $42.10 \pm 3.71$ & $0.9787 \pm 0.0229$ & $0.0072 \pm 0.0075$ & 0.199 & 5.07 & 0.91 \\
WIRE & $35.28 \pm 3.71$ & $0.9238 \pm 0.0291$ & $0.0640 \pm 0.0279$ & $37.68 \pm 3.65$ & $0.9555 \pm 0.0340$ & $0.0256 \pm 0.0149$ & 0.133 & 7.21 & 0.97 \\
SL$^2$A-INR & $38.92 \pm 3.07$ & $0.9612 \pm 0.0106$ & $0.0265 \pm 0.0101$ & $41.51 \pm 3.61$ & $0.9763 \pm 0.0198$ & $0.0096 \pm 0.0065$ & 0.330 & 5.58 & 0.85 \\
CAFE & $42.30 \pm 3.06$ & $0.9800 \pm 0.0064$ & $0.0074 \pm 0.0044$ & $45.73 \pm 3.00$ & $0.9896 \pm 0.0066$ & $0.0025 \pm 0.0014$ & 0.417 & 5.78 & 0.74 \\
CAFE+ & $48.20 \pm 2.87$ & $0.9941 \pm 0.0016$ & $0.0026 \pm 0.0023$ & $52.24 \pm 3.18$ & $0.9974 \pm 0.0018$ & $0.0004 \pm 0.0005$ & 0.472 & 7.14 & 0.90 \\
SASNet & $41.33 \pm 2.56$ & $0.9799 \pm 0.0040$ & $0.0081 \pm 0.0041$ & $44.35 \pm 2.75$ & $0.9905 \pm 0.0059$ & $0.0018 \pm 0.0009$ & 0.485 & 5.55 & 0.94 \\
SAFIR & $35.43 \pm 3.50$ & $0.9325 \pm 0.0246$ & $0.0945 \pm 0.0349$ & $39.00 \pm 3.91$ & $0.9635 \pm 0.0373$ & $0.0305 \pm 0.0290$ & 0.209 & 4.07 & 0.92 \\
SCONE & $39.16 \pm 3.09$ & $0.9623 \pm 0.0107$ & $0.0185 \pm 0.0108$ & $43.05 \pm 4.19$ & $0.9796 \pm 0.0236$ & $0.0053 \pm 0.0061$ & 0.193 & 7.55 & 1.63 \\
CF-INR & $48.86 \pm 2.24$ & $0.9959 \pm 0.0019$ & $0.0013 \pm 0.0008$ & $52.09 \pm 4.64$ & $0.9980 \pm 0.0024$ & $0.0008 \pm 0.0015$ & 0.796 & 13.89 & 2.17 \\
\midrule
AFINR-ReLU & $37.23 \pm 2.76$ & $0.9680 \pm 0.0087$ & $0.0240 \pm 0.0116$ & $37.44 \pm 3.53$ & $0.9684 \pm 0.0180$ & $0.0185 \pm 0.0090$ & 0.364 & 4.98 & 0.46 \\
AFINR-SIREN & $44.44 \pm 2.13$ & $0.9909 \pm 0.0017$ & $0.0024 \pm 0.0010$ & $46.51 \pm 2.98$ & $0.9939 \pm 0.0042$ & $0.0012 \pm 0.0007$ & 0.364 & 5.01 & 0.71 \\
AFINR-CAFE & $43.62 \pm 3.30$ & $0.9934 \pm 0.0021$ & $0.0035 \pm 0.0044$ & $49.40 \pm 4.36$ & $0.9958 \pm 0.0028$ & $0.0019 \pm 0.0047$ & 0.574 & 6.56 & 0.97 \\
AFINR-CAFE+ & \textbf{$52.15 \pm 1.93$} & \textbf{$0.9976 \pm 0.0005$} & \textbf{$0.0003 \pm 0.0002$} & \textbf{$55.74 \pm 2.43$} & \textbf{$0.9987 \pm 0.0009$} & \textbf{$0.0001 \pm 0.0001$} & 0.630 & 8.75 & 0.98 \\
\bottomrule
\end{tabular}
\end{table*}

Table~\ref{tab:main_comparison} shows that AFINR-CAFE+ achieves the best overall reconstruction quality on both datasets, improving its direct CAFE+ baseline by 3.95 dB on Kodak and 3.50 dB on DIV2K.

Compared with the recent CF-INR baseline, AFINR-CAFE+ improves PSNR by 3.29 dB on Kodak and 3.65 dB on DIV2K, while using fewer parameters (0.630M vs. 0.796M), less training time (8.75 vs. 13.89 min), and lower peak GPU memory
(0.98 vs. 2.17 GB). AFINR-SIREN also outperforms SCONE by 5.28 dB and 3.46 dB on Kodak and DIV2K, respectively.

% Auto-generated by generate_table2_decoder_gains_tex.m
\begin{table}[t]
\centering
\caption{Paired per-image gains obtained by incorporating AFINR into different decoders. For PSNR and SSIM, $\Delta$ denotes AFINR minus the baseline; for LPIPS, $\Delta$ denotes baseline minus AFINR, so positive values indicate improvement. Values are reported as mean $\pm$ standard deviation.}
\label{tab:decoder_gain}
\scriptsize
\setlength{\tabcolsep}{0.8pt}
\begin{tabular}{@{}llccc@{}}
\toprule
Baseline & AFINR variant & $\Delta$PSNR (dB)$\uparrow$ & $\Delta$SSIM$\uparrow$ & $\Delta$LPIPS$\uparrow$ \\
\midrule
\multicolumn{5}{c}{Kodak} \\
\midrule
PE-MLP & AFINR-ReLU & $+6.18 \pm 0.96$ & $+0.1434 \pm 0.0463$ & $+0.1680 \pm 0.0466$ \\
PE-SIREN & AFINR-SIREN & $+9.68 \pm 1.27$ & $+0.0304 \pm 0.0095$ & $+0.0234 \pm 0.0104$ \\
CAFE & AFINR-CAFE & $+1.31 \pm 4.40$ & $+0.0134 \pm 0.0053$ & $+0.0039 \pm 0.0070$ \\
CAFE+ & AFINR-CAFE+ & $+3.95 \pm 1.26$ & $+0.0035 \pm 0.0013$ & $+0.0023 \pm 0.0022$ \\
\midrule
\multicolumn{5}{c}{DIV2K} \\
\midrule
PE-MLP & AFINR-ReLU & $+4.82 \pm 1.40$ & $+0.0850 \pm 0.0463$ & $+0.0980 \pm 0.0415$ \\
PE-SIREN & AFINR-SIREN & $+11.67 \pm 2.50$ & $+0.0186 \pm 0.0100$ & $+0.0139 \pm 0.0066$ \\
CAFE & AFINR-CAFE & $+3.67 \pm 4.31$ & $+0.0062 \pm 0.0045$ & $+0.0006 \pm 0.0047$ \\
CAFE+ & AFINR-CAFE+ & $+3.50 \pm 1.14$ & $+0.0013 \pm 0.0010$ & $+0.0004 \pm 0.0005$ \\
\bottomrule
\end{tabular}
\end{table}

Table~\ref{tab:decoder_gain} shows positive mean gains for all four baseline--AFINR pairs on both datasets. The gains are particularly large for the ReLU and SIREN pairs, while the stronger CAFE and CAFE+ baselines also benefit from spatial frequency conditioning. The CAFE pair exhibits larger variation across images, whereas the CAFE+ gains are more stable.

\begin{figure*}[ht!]
    \centering
    \includegraphics[width=0.98\textwidth]{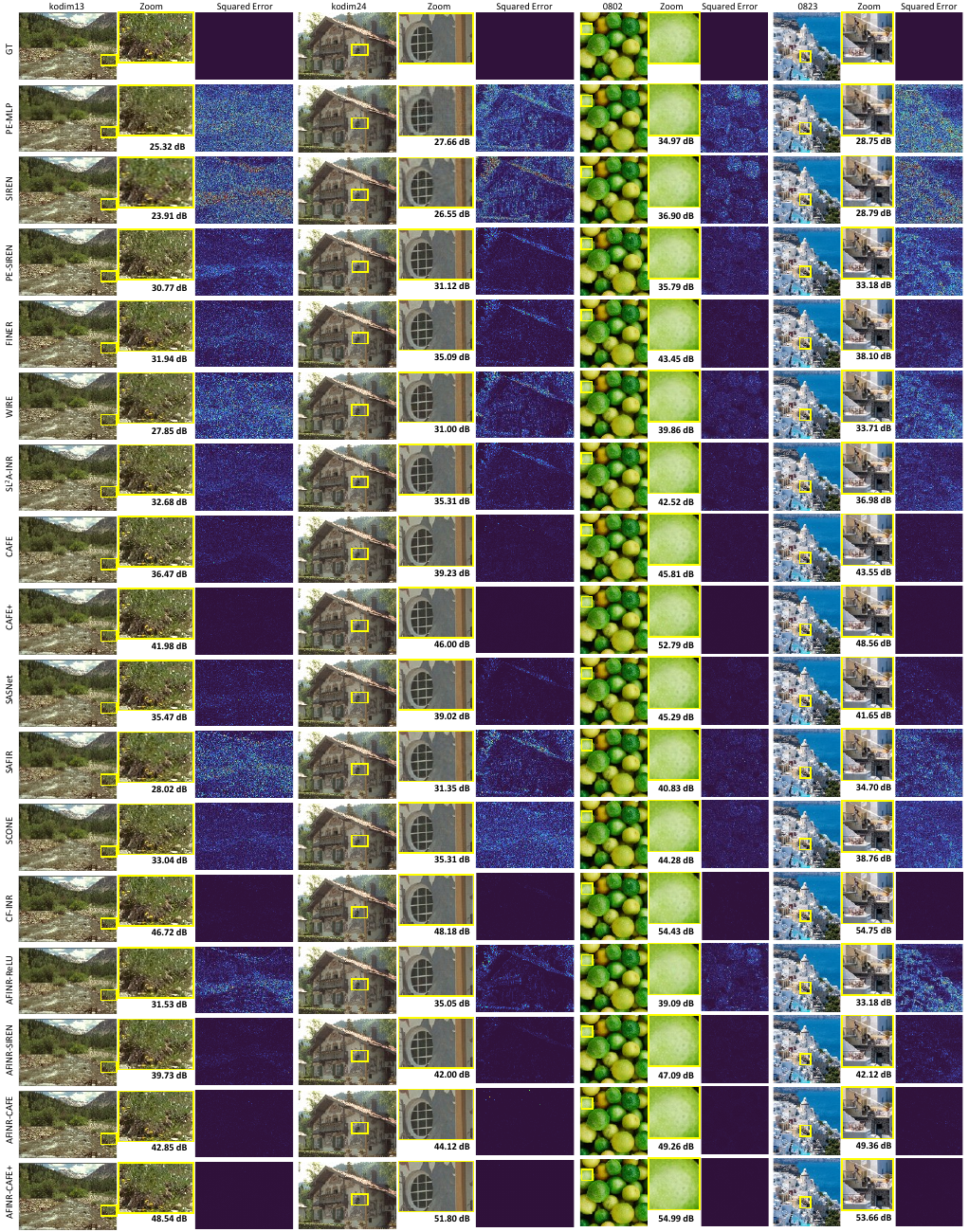}
    \caption{Qualitative comparison on representative images from Kodak and DIV2K. From left to right, the four examples correspond to \texttt{kodim13}, \texttt{kodim24}, \texttt{0802}, and \texttt{0823}. For each example, the full reconstruction, a magnified local region, and the corresponding squared-error map are presented. The rows contain the ground truth followed by all methods reported in Table~\ref{tab:main_comparison}. The same region of interest is used for all methods on each image, and the squared-error maps share the same display range within each dataset.}
    \label{fig:qualitative_comparison}
\end{figure*}

Fig.~\ref{fig:qualitative_comparison} provides qualitative comparisons on representative images from Kodak and DIV2K. The four examples include different types of image content, including natural textures (\texttt{kodim13}), architectural structures with clear boundaries (\texttt{kodim24}), repetitive fine patterns (\texttt{0802}), and urban scenes with complex structures (\texttt{0823}). The selected local regions contain both detailed patterns and structural transitions, providing examples with different types of local structures. As shown in the magnified regions and squared-error maps, the AFINR variants generally produce clearer local structures and lower visible errors compared with their corresponding baselines.

\subsection{Ablation Studies}
\label{sec:ablation}
\subsubsection{Global versus Spatial Frequency Conditioning}
\label{sec:global_spatial}

To examine whether a globally shared frequency response is sufficient, we additionally evaluate a Global-$\alpha$ variant, in which the spatial control field is replaced by a single learnable scalar, $\alpha(\mathbf{x})=\alpha_g$. The variant uses the same sampling-compatible Fourier dictionary and structured frequency-response formulation as AFINR, together with a SIREN decoder. PE-SIREN provides the unmodulated Fourier reference, while AFINR-SIREN uses the DCT-parameterized spatial control field.

\begin{table}[t]
\centering
\caption{Comparison of unmodulated, globally shared, and spatially varying frequency conditioning with SIREN-based decoders. Results are reported as mean $\pm$ standard deviation across test images.}
\label{tab:global_spatial}
\scriptsize
\setlength{\tabcolsep}{3.5pt}
\begin{tabular}{@{}lccc@{}}
\toprule
Method & PSNR (dB)$\uparrow$ & SSIM$\uparrow$ & LPIPS$\downarrow$ \\
\midrule

\multicolumn{4}{c}{Kodak} \\
\midrule

PE-SIREN
& $34.76 \pm 2.18$
& $0.9605 \pm 0.0103$
& $0.0259 \pm 0.0111$ \\

Global-$\alpha$ SIREN
& $34.22 \pm 2.28$
& $0.9465 \pm 0.0162$
& $0.0265 \pm 0.0270$ \\

AFINR-SIREN
& $\mathbf{44.44 \pm 2.13}$
& $\mathbf{0.9909 \pm 0.0017}$
& $\mathbf{0.0024 \pm 0.0010}$ \\

\midrule
\multicolumn{4}{c}{DIV2K} \\
\midrule

PE-SIREN
& $34.84 \pm 1.79$
& $0.9753 \pm 0.0139$
& $0.0151 \pm 0.0069$ \\

Global-$\alpha$ SIREN
& $34.23 \pm 1.98$
& $0.9626 \pm 0.0287$
& $0.0156 \pm 0.0360$ \\

AFINR-SIREN
& $\mathbf{46.51 \pm 2.98}$
& $\mathbf{0.9939 \pm 0.0042}$
& $\mathbf{0.0012 \pm 0.0007}$ \\

\bottomrule
\end{tabular}
\end{table}

As shown in Table~\ref{tab:global_spatial}, the Global-$\alpha$ variant remains close to, but slightly below, PE-SIREN, indicating that a single response shared across the image provides no clear advantage over the unmodulated Fourier representation. In contrast, AFINR-SIREN outperforms Global-$\alpha$ by 10.22 dB on Kodak and 12.28 dB on DIV2K, with consistent
improvements in SSIM and LPIPS. Since Global-$\alpha$ uses a single control variable whereas AFINR-SIREN employs a DCT-parameterized spatial field, the effects of field parameterization and capacity are examined in the following
ablations.

\subsubsection{Adaptive-Field Parameterization}

\begin{table}[t]
\centering
\caption{Comparison of adaptive-field parameterizations with different
decoders. Results are reported as mean $\pm$ standard deviation
across test images, with parameter counts reported at
$512\times512$ resolution.}
\label{tab:alpha_parameterization}
\scriptsize
\setlength{\tabcolsep}{2.3pt}
\begin{tabular}{llccc}
\toprule
Decoder & Parameterization &
Kodak PSNR (dB)$\uparrow$ &
DIV2K PSNR (dB)$\uparrow$ &
Params (M)$\downarrow$ \\
\midrule
ReLU  & Grid     & $41.77 \pm 2.09$ & $40.00 \pm 3.15$ & 0.469 \\
      & Hash-1   & $28.76 \pm 3.35$ & $29.05 \pm 3.52$ & 0.223 \\
      & Hash-5   & $28.22 \pm 3.37$ & $28.63 \pm 3.67$ & 0.216 \\
      & DCT-60\% & $37.23 \pm 2.76$ & $37.44 \pm 3.53$ & 0.364 \\
\midrule
SIREN & Grid     & $48.89 \pm 1.93$ & $49.38 \pm 2.54$ & 0.469 \\
      & Hash-1   & $38.71 \pm 3.00$ & $41.97 \pm 3.51$ & 0.223 \\
      & Hash-5   & $39.36 \pm 3.13$ & $42.20 \pm 3.38$ & 0.216 \\
      & DCT-60\% & $44.44 \pm 2.13$ & $46.51 \pm 2.98$ & 0.364 \\
\midrule
CAFE+ & Grid     & $55.91 \pm 1.68$ & $58.70 \pm 2.21$ & 0.734 \\
      & Hash-1   & $47.33 \pm 2.48$ & $50.98 \pm 2.84$ & 0.489 \\
      & Hash-5   & $47.37 \pm 2.18$ & $51.04 \pm 2.59$ & 0.482 \\
      & DCT-60\% & $52.15 \pm 1.93$ & $55.74 \pm 2.43$ & 0.630 \\
\bottomrule
\end{tabular}
\end{table}

The most direct implementation of the proposed adaptive field is a dense learnable grid having the same spatial resolution as the target image. Specifically, for an image of size $H\times W$, we associate one learnable value with each spatial location, resulting in an $H\times W$ adaptive field. At the $512\times512$ resolution used for parameter-count reporting, this representation alone introduces $262{,}144$ learnable parameters. As shown in Table~\ref{tab:alpha_parameterization}, this dense parameterization provides strong reconstruction accuracy, but its parameter cost increases directly with image resolution.

This observation motivates us to seek a more compact representation of the adaptive field. A natural alternative is hash encoding, which has been widely used to reduce the storage cost of spatially indexed features. We therefore consider two hash-based variants. Hash-1 uses a single-level hash field with a short-side logical resolution of 256 and a hash table containing $2^{14}$ scalar entries. Hash-5 adopts a five-level representation with resolutions $\{16,32,64,128,256\}$, two features per level, and a hash-table bitwidth of 10, followed by a lightweight $10$--$48$--$1$ decoder for predicting $\alpha(\mathbf{x})$. These two designs substantially reduce the number of parameters used to represent the adaptive field.

The reduction in parameter count, however, is accompanied by a clear loss in reconstruction accuracy. For example, with the SIREN decoder, the dense-grid representation achieves $48.89$ dB and $49.38$ dB on Kodak and DIV2K, respectively, whereas Hash-1 obtains $38.71$/$41.97$ dB and Hash-5 obtains $39.36$/$42.20$ dB. A similar degradation is observed with the ReLU and CAFE+ decoders. Under the tested parameter budgets, the hash-based variants provide substantially lower reconstruction accuracy than the dense and DCT parameterizations.

We therefore turn to a DCT parameterization. Instead of storing the adaptive field directly in the spatial domain, we represent it by DCT coefficients and reconstruct $\alpha(\mathbf{x})$ through an inverse DCT. This representation retains the global spatial organization of the dense field while allowing its parameter count to be controlled simply by truncating the retained coefficients. With all DCT coefficients retained, the DCT parameterization has the same representational capacity as the dense grid and achieves similar performance at the same parameter count, as shown in Table~\ref{tab:dct_ratio_ablation}. The small performance difference is mainly due to optimization or numerical effects.

In the default model, we retain $60\%$ of the DCT coefficients. At $512\times512$, DCT-60\% substantially reduces the parameter count relative to the dense grid while retaining considerably higher reconstruction accuracy than the hash-based variants. We therefore adopt DCT-60\% as a practical quality--compactness operating point, and examine the effect of the retention ratio in the following subsection.

\subsubsection{DCT Coefficient Ratio}
\label{sec:ablation_dct_ratio}
\begin{table}[t]
\centering
\caption{Ablation of the retained DCT coefficient ratio using a fixed SIREN decoder and training protocol. Results are reported as mean $\pm$ standard deviation across test images, with parameter counts reported at $512\times512$. Bold indicates the default quality--compactness operating point.}
\label{tab:dct_ratio_ablation}
\scriptsize
\setlength{\tabcolsep}{5.5pt}
\begin{tabular}{lccc}
\toprule
Retained ratio & Kodak PSNR (dB)$\uparrow$ & DIV2K PSNR (dB)$\uparrow$ & Params (M)$\downarrow$ \\
\midrule
DCT-50\% & $43.62 \pm 2.21$ & $45.80 \pm 3.11$ & 0.338 \\
\textbf{DCT-60\%} & $44.44 \pm 2.13$ & $46.51 \pm 2.98$ & 0.364 \\
DCT-70\% & $45.41 \pm 2.19$ & $47.21 \pm 2.72$ & 0.391 \\
DCT-80\% & $46.63 \pm 2.32$ & $47.89 \pm 2.79$ & 0.417 \\
DCT-90\% & $47.74 \pm 2.00$ & $48.69 \pm 2.72$ & 0.443 \\
DCT-100\% & $48.66 \pm 2.14$ & $49.05 \pm 2.81$ & 0.469 \\
\bottomrule
\end{tabular}
\end{table}

We investigate the effect of the retained DCT coefficient ratio with the SIREN decoder and optimization settings fixed. As shown in Table~\ref{tab:dct_ratio_ablation}, reconstruction accuracy increases consistently as more DCT coefficients are retained. On Kodak, the PSNR increases from 43.62 dB at 50\% retention to 48.66 dB at 100\%, while on DIV2K it increases from 45.80 to 49.05 dB. This trend reflects the increased degrees of freedom available to the spatial control field as the retention
ratio increases.

The 100\% setting also provides a direct comparison with the dense-grid parameterization in Table~\ref{tab:alpha_parameterization}. With the SIREN decoder, DCT-100\% and the dense grid use the same 0.469M parameters and achieve 48.66 versus 48.89 dB on Kodak and 49.05 versus 49.38 dB on DIV2K, respectively. Since the full orthonormal DCT spans the same $H\times W$ spatial-field space as the dense representation, their similar performance is consistent with their equivalent representational capacity.

We use DCT-60\% as the default compact configuration. At $512\times512$, it reduces the total parameter count from $0.469$M to $0.364$M, corresponding to a reduction of approximately $22.4\%$, while retaining substantially higher accuracy than the hash-based parameterizations. We therefore select DCT-60\% as a practical quality--compactness operating point rather than an
accuracy-maximizing setting.

\subsubsection{Sensitivity to Frequency-Response Parameters}

\begin{table}[h]
\centering
\caption{Sensitivity to the frequency-response parameters $B$ and
$\kappa$ using AFINR-SIREN with DCT-60\%. Results are reported as
mean $\pm$ standard deviation across test images. Bold parameter
values indicate the default settings.}
\label{tab:response_sensitivity}
\scriptsize
\setlength{\tabcolsep}{6.5pt}

\begin{tabular}{llcc}
\toprule
Parameter & Value & Kodak PSNR (dB)$\uparrow$ & DIV2K PSNR (dB)$\uparrow$ \\
\midrule
$B$ ($\kappa=12$) & 0.30 & $43.77 \pm 2.31$ & $45.20 \pm 2.98$ \\
                   & 0.40 & $44.25 \pm 2.35$ & $45.76 \pm 2.93$ \\
                   & \textbf{0.50} & $44.44 \pm 2.13$ & $46.51 \pm 2.98$ \\
                   & 0.60 & $44.39 \pm 2.50$ & $46.54 \pm 3.02$ \\
                   & 0.70 & $44.02 \pm 2.61$ & $46.40 \pm 3.41$ \\
\midrule
$\kappa$ ($B=0.50$) & 4  & $43.35 \pm 2.32$ & $45.09 \pm 3.24$ \\
                     & 8  & $44.35 \pm 2.27$ & $46.26 \pm 3.19$ \\
                     & \textbf{12} & $44.44 \pm 2.13$ & $46.51 \pm 2.98$ \\
                     & 16 & $44.26 \pm 2.46$ & $46.46 \pm 3.03$ \\
                     & 20 & $44.34 \pm 2.29$ & $46.43 \pm 3.03$ \\
\bottomrule
\end{tabular}
\end{table}

We further examine the sensitivity of AFINR to the two parameters that control the shape of the frequency response, namely the response width $B$ and the transition sharpness $\kappa$. To isolate their effects, we fix the decoder to SIREN and use DCT-60\% for the adaptive field throughout this experiment. When varying one parameter, the other is kept at its default value ($B=0.50$ and $\kappa=12$).

As shown in Table~\ref{tab:response_sensitivity}, the reconstruction performance is relatively stable over a moderate range of $B$. On Kodak, the PSNR increases from $43.77$ dB at $B=0.30$ to $44.44$ dB at $B=0.50$, and remains close to this level for $B=0.60$. A similar behavior is observed on DIV2K, where $B=0.50$ and $B=0.60$ yield $46.51$ dB and $46.54$ dB, respectively. The degradation at smaller $B$ indicates that an overly narrow response restricts the range of frequency components available at each spatial location, whereas further increasing the response width provides little additional benefit.

The results for $\kappa$ exhibit a similar trend. A relatively small value of $\kappa=4$ reduces the PSNR to $43.35$ dB on Kodak and $45.09$ dB on DIV2K. Increasing $\kappa$ to 8 substantially improves the reconstruction, while the results remain stable between $\kappa=8$ and $\kappa=20$. The default value $\kappa=12$ achieves $44.44$ dB on Kodak and $46.51$ dB on DIV2K, with only minor variations observed for larger values. This behavior suggests that a sufficiently sharp, but still smooth, frequency transition is beneficial, while making the transition considerably sharper does not lead to further improvement.

Based on these results, we use $B=0.50$ and $\kappa=12$ in all main experiments. The stable performance around these settings also indicates that AFINR is not overly sensitive to the frequency-response parameters.

\subsection{Further Analysis}
\subsubsection{Parameter Budget Analysis}
\label{sec:parameter_analysis}

\begin{table}[t]
\centering
\caption{Comparison under comparable parameter budgets. Parameter
counts are reported as equivalent values at $512\times512$
resolution.}
\label{tab:capacity_equivalent}
\scriptsize
\setlength{\tabcolsep}{1.5pt}
\begin{tabular}{lcccc}
\toprule
Method & Params (M) & PSNR (dB)$\uparrow$ & SSIM$\uparrow$ & LPIPS$\downarrow$ \\
\midrule
\multicolumn{5}{c}{Kodak} \\
\midrule PE-SIREN & 0.2071 & $34.76 \pm 2.18$ & $0.9605 \pm 0.0103$ & $0.0259 \pm 0.0111$ \\
Large PE-SIREN & 0.3649 & $36.15 \pm 2.16$ & $0.9798 \pm 0.0053$ & $0.0161 \pm 0.0078$ \\
AFINR-SIREN & 0.3645 & $44.44 \pm 2.13$ & $0.9909 \pm 0.0017$ & $0.0024 \pm 0.0010$ \\
\midrule
\multicolumn{5}{c}{DIV2K} \\
\midrule
PE-SIREN & 0.2071 & $34.84 \pm 1.79$ & $0.9753 \pm 0.0139$ & $0.0151 \pm 0.0069$ \\
Large PE-SIREN & 0.3649 & $35.45 \pm 1.61$ & $0.9849 \pm 0.0057$ & $0.0117 \pm 0.0068$ \\
AFINR-SIREN & 0.3645 & $46.51 \pm 2.98$ & $0.9939 \pm 0.0042$
& $0.0012 \pm 0.0007$ \\
\bottomrule
\end{tabular}
\end{table}

To examine whether the improvement of AFINR mainly comes from the additional adaptive-field parameters, we compare AFINR-SIREN with a widened PE-SIREN under a comparable parameter budget. The PE-SIREN decoder is enlarged by increasing its hidden width while keeping the Fourier encoding and SIREN structure unchanged. Since the adaptive-field size depends on spatial resolution, all parameter counts are reported as equivalent values at $512\times512$.

As shown in Table~\ref{tab:capacity_equivalent}, increasing PE-SIREN from 0.207M to 0.365M parameters yields only 1.39 dB and 0.61 dB PSNR gains on Kodak and DIV2K, respectively. AFINR-SIREN achieves substantially higher reconstruction accuracy under a comparable parameter budget, indicating that increased decoder capacity alone cannot explain the improvement and that spatial frequency conditioning contributes essentially to the gain.

\subsubsection{Optimization Behavior}
\begin{figure}[t]
    \centering
    \includegraphics[width=0.45\textwidth]{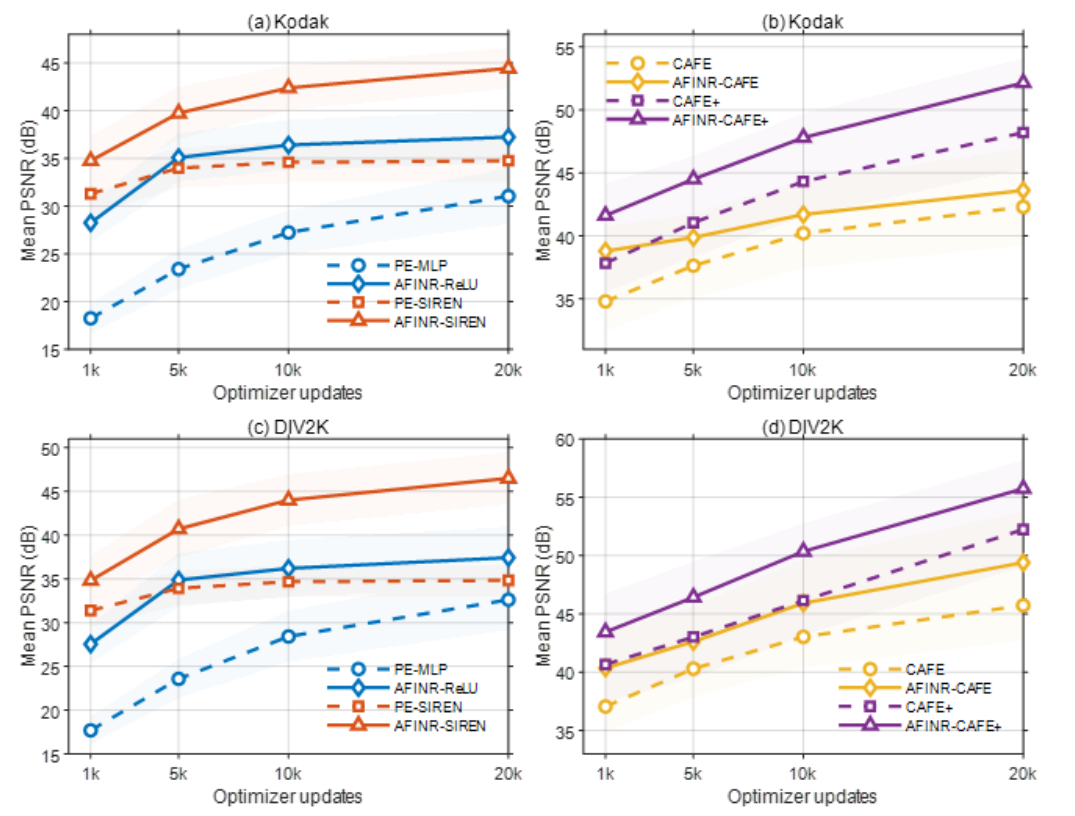}
    \caption{Optimization behavior of matched baseline--AFINR pairs. (a,b) Kodak and (c,d) DIV2K; (a,c) show the ReLU and SIREN pairs, while (b,d) show the CAFE and CAFE+ pairs. Curves report mean PSNR at 1k, 5k, 10k, and 20k updates, with shaded regions denoting one standard deviation.}
    \label{fig:optimization_behavior}
\end{figure}

Fig.~\ref{fig:optimization_behavior} compares the optimization trajectories of the four matched baseline--AFINR pairs. AFINR achieves higher mean PSNR than the corresponding baseline at every recorded checkpoint on both datasets, including the stronger CAFE and CAFE+ configurations. This shows that the advantage is present throughout optimization under a fixed update budget. 

\subsubsection{Learned Spatial Frequency Adaptation}
\label{sec:frequency_adaptation}
\begin{figure}[t]
    \centering
    \includegraphics[width=0.45\textwidth]{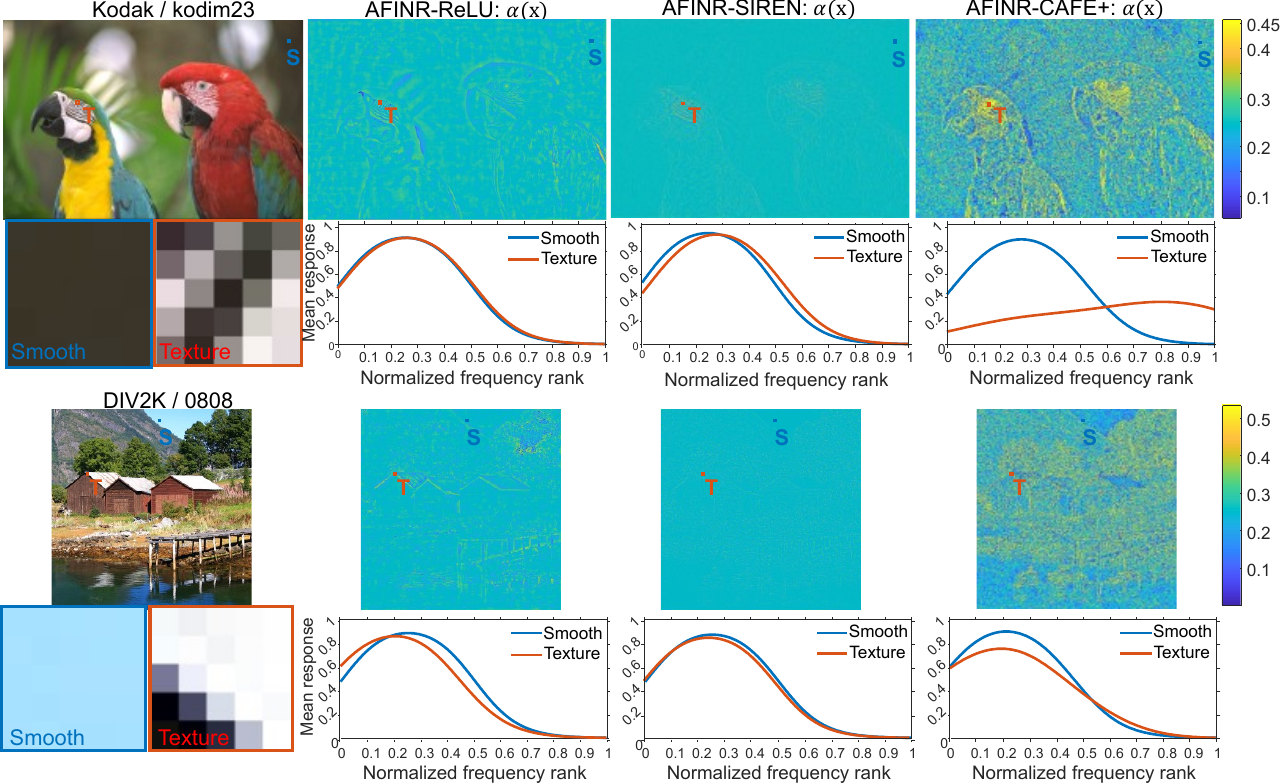}
    \caption{Learned spatial frequency adaptation on Kodak \textit{kodim23} (top two rows) and DIV2K \textit{0808} (bottom two rows). For each image, the first row shows the ground truth and learned $\alpha(\mathbf{x})$ fields of AFINR-ReLU, AFINR-SIREN, and AFINR-CAFE+, and the second row shows representative smooth and textured regions with their frequency responses. Blue and orange denote the smooth and textured regions, respectively.}
    \label{fig:frequency_adaptation}
\end{figure}

To further examine the behavior of the proposed adaptive frequency modulation, Fig.~\ref{fig:frequency_adaptation} visualizes the learned spatial field $\alpha(\mathbf{x})$ and the induced frequency responses for representative Kodak and DIV2K images. For each example, a relatively smooth region and a textured region with richer local spatial variations are selected for comparison. The learned $\alpha(\mathbf{x})$ exhibits clear spatial variation, and the corresponding regions induce different frequency responses. The degree of separation varies across decoder backbones: it is relatively moderate for AFINR-ReLU and AFINR-SIREN, while AFINR-CAFE+ shows more pronounced region-dependent modulation. These observations indicate that the adaptive field does not behave as a globally fixed frequency filter, but instead modulates the explicit frequency representation according to spatially varying image content.

\section{Discussion}
\label{sec:discussion}
AFINR consistently improves different coordinate decoders, including ReLU, SIREN, CAFE, and CAFE+, indicating that spatial frequency conditioning provides complementary benefits to different INR architectures. Its effectiveness can be understood from two aspects: the sampling-compatible Fourier basis restricts the explicit frequencies to those supported by the observations, while the spatial control field allows different regions to use the admitted spectrum differently.

The ablation studies further clarify the source of the improvement. The Global-$\alpha$ variant remains close to the unmodulated PE-SIREN baseline, whereas spatially varying AFINR achieves substantially higher accuracy, showing that spatial variation of the frequency response is important. Together with the comparable parameter-budget experiment, the results also indicate that the gain cannot be explained solely by increased model capacity.

The adaptive field introduces an accuracy--compactness tradeoff. Dense-grid parameterization provides high representation capacity but scales directly with image resolution, whereas the tested hash-based parameterizations are more compact but less accurate. The DCT parameterization provides an intermediate solution by preserving the spatial organization of the field while explicitly controlling its degrees of freedom through coefficient retention. DCT-60\% is therefore used as a compact operating point rather than an accuracy-maximizing setting.

AFINR also differs from existing spatially adaptive Fourier methods in the form of modulation. SASNet and SCONE use spatially varying masks to adjust neurons or Fourier bases across locations, whereas AFINR uses a scalar spatial control field to generate a structured response along the ordered frequency axis. The Fourier dictionary therefore remains fixed while its effective contribution varies spatially, providing an interpretable relation between location and frequency use.

AFINR has several limitations. The adaptive field increases model size and computational cost compared with fixed Fourier representations, and the current response is restricted to a low-dimensional parametric form controlled by $B$, $\kappa$, and $\alpha(\mathbf{x})$. Future work can investigate more flexible response functions and extend the framework to inverse problems such as sparse-view CT, super-resolution, and light-field reconstruction.

\section{Conclusion}
\label{sec:conclusion}

This work addresses the spatially uniform use of Fourier features in INRs, which does not account for spatial variations in signal frequency content. We proposed AFINR, which uses a learnable spatial control field to modulate the admitted Fourier components and parameterizes this field in the DCT domain for compact representation. Experiments on Kodak and DIV2K show consistent average improvements across ReLU, SIREN, CAFE, and CAFE+ decoders, while the ablation studies characterize the accuracy--compactness tradeoff of the adaptive field. The results support spatially adaptive frequency conditioning as an effective strategy for Fourier-encoded INRs.

\bibliography{ref.bib}
\bibliographystyle{IEEEtran}

\end{document}